\documentclass{article}

\usepackage[accepted]{ai4science}

\usepackage[utf8]{inputenc} 
\usepackage[T1]{fontenc}    
\usepackage{hyperref}       
\usepackage{url}            
\usepackage{booktabs}       
\usepackage{amsfonts}       
\usepackage{nicefrac}       
\usepackage{microtype}      
\usepackage{lipsum}
\usepackage{graphicx}
\usepackage{amsmath}
\usepackage{color}
\usepackage[misc]{ifsym}

\icmltitlerunning{From Kepler to Newton: Explainable AI for Science}

\begin{document}

\twocolumn[
\icmltitle{From Kepler to Newton: Explainable AI for Science}


\begin{icmlauthorlist}
\icmlauthor{Zelong Li}{rutgers}
\icmlauthor{Jianchao Ji}{rutgers}
\icmlauthor{Yongfeng Zhang}{rutgers}
\end{icmlauthorlist}

\icmlaffiliation{rutgers}{Department of Computer Science, Rutgers University, New Brunswick, NJ 08854, USA}

\icmlcorrespondingauthor{Yongfeng Zhang}{yongfeng.zhang@rutgers.edu}

\icmlkeywords{Explainable AI, Science Discovery, Scientific Method, Technological Singularity, Human and Nature}

\vskip 0.3in
]



\printAffiliationsAndNotice{}  

\begin{abstract}
The Observation --- Hypothesis --- Prediction --- Experimentation loop paradigm for scientific research has been practiced by researchers for years towards scientific discoveries. However, with data explosion in both mega-scale and milli-scale research, it has been sometimes very difficult to manually analyze the data and propose new hypotheses to drive the cycle for scientific discovery.

In this paper, we discuss the role of Explainable AI in scientific discovery process by demonstrating an Explainable AI-based paradigm for science discovery. The key is to use Explainable AI to help derive data or model interpretations, hypotheses, as well as scientific discoveries or insights. We show how computational and data-intensive methodology---together with experimental and theoretical methodology---can be seamlessly integrated for scientific research. To demonstrate the AI-based science discovery process, and to pay our respect to some of the greatest minds in human history, we show how Kepler's laws of planetary motion and Newton's law of universal gravitation can be rediscovered by (Explainable) AI based on Tycho Brahe's astronomical observation data, whose works were leading the scientific revolution in the 16-17th century. This work also highlights the important role of Explainable AI (as compared to Blackbox AI) in science discovery to help humans prevent or better prepare for the possible technological singularity that may happen in the future, since science is not only about the know how, but also the know why.



\end{abstract}


\vspace{-4ex}

\section{Introduction}\label{sec:introduction}

\begin{figure}[t]
    \centering
    \includegraphics[width=0.5\textwidth]{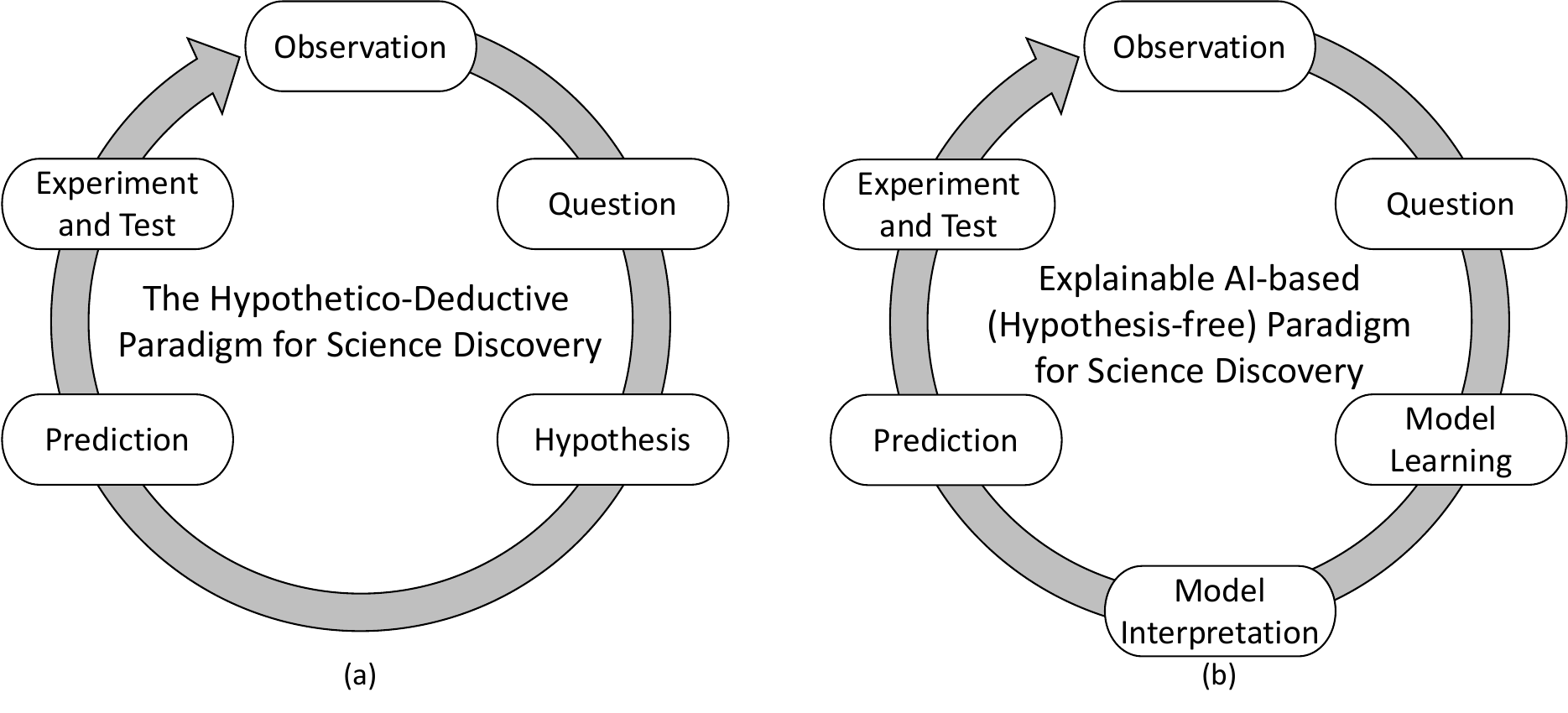}
    \vspace{-4ex}
    \caption{The Hypothetico-Deductive paradigm for science discovery and the Explainable AI-based (hypothesis-free) paradigm for science discovery. The new paradigm uses Explainable AI to generate verifiable hypothesis.}
    \label{figure:paradigm}
    \vspace{-3ex}
\end{figure}

A frequently used paradigm for 
scientific research
is the Hypothetico-Deductive paradigm (Figure \ref{figure:paradigm}(a)), which has been practiced by researchers for years \cite{godfrey2009theory,nola2014theories,hey2009fourth}. In this paradigm, researchers first make observations which is usually a data collection process, and then raise a question. To get answers to the question, researchers will then propose a hypothesis as a possible explanation to the observation, usually through an abductive reasoning process. The hypothesis may come in the form of a theory, a model, an equation, an algorithm, or any other form depending on the research problem and research area. The hypothesis is used to make verifiable predictions, and then experimental tests or data analyses are conducted to verify or falsify the hypothesis. The above process may repeat as a loop, i.e., if the hypothesis is falsified, we may need to make new observations, propose new hypotheses, and even ask a new question. 

\begin{figure}[t]
    \centering
    \includegraphics[width=0.5\textwidth]{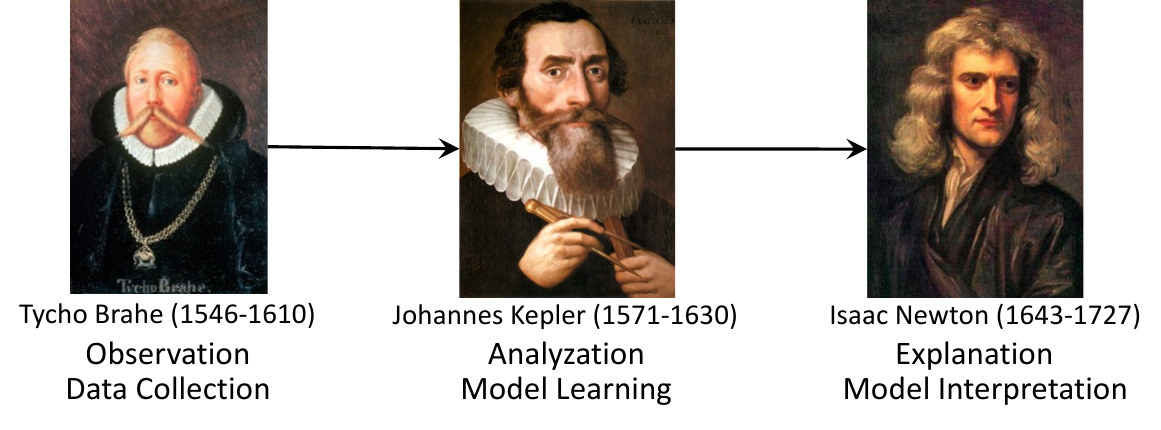}
    \vspace{-4ex}
    \caption{Tycho Brahe, Johannes Kepler, Isaac Newton and their roles in the science discovery process.}
    \label{figure:brahe_kepler_newton}
    \vspace{-3ex}
\end{figure}

An excellent example of science discovery is the works of Tycho Brahe, Johannes Kepler and Isaac Newton (Figure \ref{figure:brahe_kepler_newton}), who are some of the greatest minds in human history and their work were leading the scientific revolution in the 16-17th century. Tycho Brahe was an astronomer known for his accurate and comprehensive astronomical observations. During his career in the 16th century, though as a naked-eye astronomer, his observations of the planets orbiting the Sun were so accurate that it became possible for later researchers to build insightful discoveries based on his observational data. One notable name, of course, is Johannes Kepler, who discovered what was later known as the Kepler's laws of planetary motion. During the science discovery process, Kepler hypothesized that the orbit of a planet is an ellipse with the Sun at one of the two foci, and he was able to fit the orbital equation of Mars based on Tycho's observational data. Finally, the equation turns out to be surprisingly accurate in predicting the future position of planets, which verifies his first law of planetary motion. Later, Kepler further discovered the second and third laws through his insightful analyses of the data. Isaac Newton, one of the most notable figures in the human history of science, was not only interested in how planets orbit the Sun, but also why they orbit in such a way, which means that his goal is to explain the underlying mechanism of planetary motions. In conquest of this goal, he made several innovate discoveries which are later known as the Newton's law of universal gravitation and the Newton's laws of motion.


It is very interesting to see that Tycho, Kepler and Newton---though their works span over a hundred years of history---actually play different but closely related roles in the science discovery process, which are observation, analyzation, and explanation. Tycho's key contribution is on observation and his accurate data lays foundation for insightful analyses and innovative discoveries in the future. Kepler analyzed the data and discovered meaningful patterns hidden in the data. Finally, Newton examined the underlying mechanism of such patterns and provided insightful explanations to show why planets move in such patterns rather than other patterns. Using more computer science language, Tycho's work is on data collection, Kepler's work is on model learning, i.e., he manually (instead of using modern computers) fit the data and learned predictive models based on the data, and finally, Newton's work is on model interpretation, i.e., he (also manually) provided conceptual and mathematical explanations for Kepler's results and Kepler's laws can be naturally derived from Newton's laws.

In modern scientific research, with the help of various mechanical, electrical and biological equipment, many components of the research pipeline have been automated. The most notable component is observation and data collection---modern equipment such as telescopes, sensors and colliders automatically and continuously collect data to support research and discoveries, and such observational data usually comes in massive scale. For example, the Hubble Space Telescope (HST) generates up to 150 GB of spatial data per week \cite{eldawy2015era}, and the Large Hadron Collider (LHC) experiments produce about 90 PB of data per year \cite{naim2020pushing}. Such abundant and accurate observational data helps to push the frontier of scientific research, but it also brings great challenges to process the data and build insightful hypotheses from the data. However, building insightful hypotheses is vitally important to drive the research cycle for new scientific discoveries.


To indicate how the above challenges can be alleviated with the help of modern AI and machine learning technologies, we show an Explainable AI-based hypothesis-free paradigm for science discovery (Figure \ref{figure:paradigm}(b)). The key is to replace manual hypothesis development with an AI-driven model learning and model interpretation process. More specifically, the model learning component adopts black-box AI tools such as deep learning for data analysis, data augmentation and building accurate prediction models, while the model interpretation component adopts Explainable AI tools such as symbolic regression to translate the black-box model into human-understandable forms for understanding the scientific meanings and deriving scientific insights. Working together, the two components turn manual hypothesis development into automatic hypothesis development, saving efforts of building insightful hypotheses from data. 

As a demonstration to the Explainable AI-based paradigm,
we show how the Kepler's laws of planetary motion and the Newton's law of universal gravitation can be rediscovered by explainable AI based on Tycho's astronomical observation data. 
At Kepler's time, there were three main hypotheses on planetary motion---the Tychonic system, the Ptolemaic system and the Copernican system. These three hypotheses can give good predictions in the short term, but diverge from the observational data in the long term. Kepler spent several years on calculations and finally rejected the three models.
Meanwhile, he proposed his elliptical orbit hypothesis and verified that it had better predictions than previous models. This process follows the observation--hypothesis--prediction paradigm of science discovery. Our experiments imagine that AI and machine learning techniques existed in Kepler's time and show how Kepler would have been able to derive his first and second laws using Explainable AI-based science discovery based on the observational data of Mars, without manually making hypotheses.
Besides, Explainable AI not only helps to find the first and second laws of planetary motion, but also helps
the discovery process of Kepler's third law: the ratio between the square of a planet's orbital period and the cube of the length of the semi-major axis of its orbit, is a constant for all planets.
It seems impossible to find the third law since that needs the observational data of other planets, but our experiment will show that by only using Mars data, Explainable AI is able to find the numerical relationship between the angular speed and the distance from the Sun with very high accuracy, which provides a clear direction towards the discovery of the third law.


Throughout Kepler's research career in the first 30 years of the 17th century, especially after his three laws of planetary motion have been discovered, 
Kepler has been constantly seeking for a kinetic explanation for the laws. He tried to explain the laws based on magnetic force, which from modern perspective turned out to be incorrect \cite{Astronomia}. However, it demonstrates humans' eager for explanations so as to not only know how but also know why. This is also the reason why we emphasize the importance of Explainable AI in the science discovery process.

History assigned the duty of finding the explanation to Issac Newton. If we would like to dig out the secret behind the motion of planets, we need to leverage the concepts of force and acceleration. During Kepler's time, scientists have established the concept of force, but they still do not know the exact function of force. They thought force was proportional to distance and thus speed. 
Newton's greatness lies in that he creatively linked the relationship between force and acceleration (instead of speed) and proposed the inverse square law of force and thus acceleration. 
Based on these concepts, Kepler's laws can be naturally derived from Newton's laws based on mathematical deviations, thus providing an explanation for the underlying mechanism of Kepler's laws of planetary motion \cite{newton2020principia}.

On the other hand, as Newton stated in his groundbreaking work the \textit{Principia}, he considered forces from a mathematical point of view, not a physical view, and thus taking an instrumentalist view of his methods \cite{newton2020principia}. As a result, it is the job of the readers to assign ``meanings'' to the many variables such as the force in his mathematical framework. Though contemporary scientists believe that Newton must have his own insightful understanding of the meanings of the variables, and his statement was just making room for flexibility to accommodate different views of his readers, his cautiousness inspires us to think about what is the exact role of (Explainable) AI in science discovery. In the context of Explainable AI-based science discovery, the AI machines could indeed be able to learn black-box neural models for prediction and learn symbolic equations to explain the predictions,
however, machines do not possess ``meaning'' of the variables or the combination of variables in the equations. For machines, the variables are just symbols used for data exploration, data fitting and prediction, while it is the job of humans to assign meanings to the variables and to build understandings of the universe based on the discoveries made by AI machines.\footnote{In this discuss we limit ourselves to the sense of ``meaning'' in terms of human's perspective. It is possible that machines would build their own internal ``meaning'' of the variables and calculations that is not understandable to humans, but that is beyond the scope of discussion in this work.} The role of (Explainable) AI in the science discovery process is to produce valid hypotheses or to narrow down the search space of hypotheses so as to speed up the discovery, but the role of assigning ``meanings'' to the discoveries is the job of humans (especially domain experts) which cannot be replaced by AI.
In the experiments, we will demonstrate the importance of human during science discovery by showing how the representation of force is discovered by AI when explaining the elliptical orbit and how human needs to intervene so as to assign an appropriate meaning to force.




This work also highlights the importance of Explainable AI (as compared to black-box AI) in science discovery. In particular, Explainable AI helps human beings to prevent or better prepare for the possible technological singularity (or simply singularity) that may happen in the future.
The possibility of technology advancements leading to a singularity has been discussed by public figures from many fields such as John von Neumann \cite{ulam1958tribute}, Irving John Good \cite{vinge1993coming} and Stephen Hawking \cite{sparkes2015top}. For example, I. J. Good speculated in 1965 that the advancement of artificial intelligence may bring about an intelligence explosion, where intelligent machines can solve problems and even build new machines using incomprehensible ways for humans and thus the intelligence of human would be left far behind \cite{good1966speculations}. 
Under the context of science discovery, if we develop and allow black-box AI to make discoveries and represent such discoveries using black-box models such as complex neural networks that are incomprehensible for humans (though these models may indeed provide accurate predictions), it may lead to the situation that machines will accumulate knowledge that are more and more incomprehensible for humans and eventually the human knowledge will be left far behind by machine's knowledge, leading to the singularity and even making machines out of control for humans. As a result, we need to make sure that AI explains its model and discoveries to humans using human understandable methods, so that humans can always keep track of the new discoveries and knowledge created by machines during the science discovery process.

In the following part of this paper, we first introduce some related work in Section \ref{sec:related_work}, and then we will use Kepler's and Newton's works as examples to demonstrate the Explainable AI-based science discovery process. More specifically, in Section \ref{sec:setting}, we will first introduce the data and Explainable AI models to be used in this work, and then we will rediscover Kepler's and Newton's laws under the Explainable AI-based paradigm in Section \ref{sec:rediscover_kepler_1} and Section \ref{sec:rediscover_newton}, respectively. 
We conclude the work together with discussions and future directions in Section \ref{sec:conclusions}.

\vspace{-1ex}
\section{Related Work}
\label{sec:related_work}

\vspace{-1ex}
\subsection{Explainable AI}

Explainability has been an important perspective to consider in many AI systems, leading to the research on Explainable AI (XAI). For example, recommender system needs to explain its recommendations or decisions to users so as to gain trust and help users make informed decisions, leading to the research on explainable recommendation \cite{zhang2020explainable,zhang2014explicit,tan2021counterfactual}; many prediction or classification algorithms need to provide explanations for the model designers to help them understand how the model works for better debugging and detecting potential bias in models \cite{shapley1953,ribeiro2016should,sundararajan2020many}. Explainable AI methods can be generally classified to model-intrinsic methods and model-agnostic methods \cite{zhang2020explainable}. For model-intrinsic methods, the decision and explanation are both produced by the same model, which means that the working mechanism of the model itself is transparent so that any decision produced by the model are naturally accompanied with explanations. The decision and explanation are usually produced concurrently in model-intrinsic methods. Notable examples of model-intrinsic methods include linear regression \cite{seber2012linear}, decision tree \cite{quinlan1986induction,quinlan2014c4} and attention mechanism \cite{zhang2014explicit,bahdanau2015neural,wiegreffe2019attention,jain2019attention}, whose explanations are regression coefficients, decision paths and attention weights, respectively. 
For model-agnostic methods, the decision model and explanation model are usually two separate models. The decision model is responsible for prediction and decision making, while the explanation model is responsible for explaining the results produced by the decision model. In model-agnostic methods, the explanations are usually produced in a post-hoc manner, i.e., the model decisions are produced first and then explanations are generated for the decisions. Notable examples for model-agnostic methods include counterfactual explanations \cite{tan2021counterfactual}, local approximations \cite{ribeiro2016should}, and Shapley values \cite{shapley1953,sundararajan2020many}.

It is worth noting that there exist explanation methods that may not be simply classified as either model-agnostic or model-intrinsic but actually in between agnostic and intrinsic because they can be implemented in either way. One such example is symbolic regression \cite{koza1992genetic,lu2016using}.
Symbolic regression is a type of regression analysis to find a function $f(x_1, x_2, ..., x_n) = y$ consisting of designated base functions that best fits the given dataset \cite{koza1992genetic}. The base functions could be basic number operations such as addition, subtraction, multiplication, division, exponentiation, logarithm, etc., or trigonometric functions such as sine, cosine, tangent, etc., or any other designated base functions. Symbolic regression can be done in an intrinsic way by directly learning the symbolic function that best regresses the data, or can be done in an agnostic/post-hoc way by first learning a black-box model such as neural network to fit the data and then using symbolic function to regress the black-box model. Symbolic regression is an NP-hard optimization problem \cite{lu2016using, towfighi2020symbolic}, but some effective and efficient heuristic methods have been developed, including genetic programming \cite{augusto2000symbolic, gustafson2005improving}, Bayesian methods \cite{jin2019bayesian}, and continuous optimization methods \cite{udrescu2020ai, udrescu2020ai2}. Besides, due to the high demand of solving symbolic regression problems in industry and research, many packages and tool-kits have been developed, such as Eureqa \cite{schmidt2009distilling} which is based on genetic programming and TuringBot \cite{2020turingBot} which is based on simulated annealing.

\vspace{-1ex}
\subsection{AI for Science Discovery}

AI for science discovery has been an important direction and is especially trending in recent years. For example, many efforts have been devoted to explore machine learning for drug discovery \cite{smalley2017ai, smith2018transforming, chan2019advancing}, material design \cite{gomes2019artificial}, and chemistry or physics problems \cite{haghighat2020deep, cranmer2020discovering, kusaba2021recreation}, though many of the works are conducted on synthesized data such as particle interaction rather than real observational data. A notable recent advance on AI for molecular biology is AlphaFold \cite{jumper2021highly}, which develops deep learning models to predict the folding structure of proteins. Most existing research on AI for science discovery focus on the AI utility instead of the AI explainability, i.e., they focus on developing advanced AI models for more accurate prediction, classification or regression of scientific data, but less effort is put on explaining the AI models or the AI-based discoveries. However, we believe that enabling AI to provide insightful explanations for science discovery is critically important, since it helps researchers to better understand the underlying mechanism of the AI models and better understand the scientific insight implied by the AI models, which is important to enhance human-beings' understanding of the AI-discovered knowledge and advance science progress in the community.





\section{Research Setting and Background}
\label{sec:setting}

\subsection{Research Data}

Modern research facilities such as advanced telescopes have been able to collect very accurate and abundant data for planets orbiting the Sun. However, to fully restore the research situation at Kepler's and Newton's time, and to show how Kepler's and Newton's laws can be rediscovered by Explainable AI based on the (limited) data and knowledge at their time, we do not use any of the modern data of planetary motion. Instead, throughout the research, we only use the data and knowledge that were available to and used by Kepler, which was mostly collected by Tycho Brahe and partly collected or refined by Kepler 400 years ago. In particular, we use the observation data of Mars orbiting the Sun by Tycho and Kepler
as summarized in Table \ref{tab:position of Mars}, which comes from Kepler's epoch-making book Astronomia Nova \cite{Astronomia}.


In Table \ref{tab:position of Mars}, the date is written in old style used by Kepler.\footnote{This old style is based on Julian calendar which was used by Kepler. To fix the calendar drift of spring equinox due to the excess leap days introduced by the Julian algorithm, a calendar reform was introduced in 1582 which slightly adjusted the number of days per year and advanced the date by 10 days: October 4, 1582 was followed by October 15, 1582 \cite{Gravissimas,cohen2000adoption}, leading to what is now known as the Gregorian calendar.} To obtain Gregorian style dates, we just need to add 10 days on top of Kepler's dates \cite{Astronomia}. The ``Mars' Angular Position'' from Sun is the Mars' longitudes in heliocentric ecliptic coordinates computed by Kepler.
The ``Sun-Mars Distance'' is in units of Astronomical Unit (AU). A note here is that at Kepler's time, humans were still unable to measure the distance between planets in miles \cite{eremenko2016kepler, wilson1972did}. Instead, they recorded distances in the ratio of Sun-Earth distance, and set the average of Sun-Earth distance as $100,000$, similar to the definition of AU. Thus, we use AU as the unit.\footnote{Actually, computing Mars' position and distance relative to Sun is one of Kepler's most genius innovations. He smartly used the fact that the Mars' period is 687 days, and thus should appear at the same position in universe once every 687 days. This makes it possible to compute the Sun-Mars distance and direction relative to the Sun-Earth distance (which is 1 AU) based on trigonometric calculations \cite{Astronomia}.} ``Difference'' is the difference between the computed and the observed Mars' positions in geocentric ecliptic longitudes. Since the average measurement error of Tycho's observations is just several arc-minutes and the largest difference is less than six arc-minutes, thus, we take the data as the top-accurate data at Tycho's and Kepler's time.





\subsection{AI and Explainable AI Models}

In this research, we 
aim to show what AI is able and unable to do in science discovery. In particular, we will highlight that learning black-box models could indeed give us accurate predictions of the physical phenomena, but may not help in advancing human understandings of the nature and universe. To really transform data into knowledge rather than just prediction tools, we not only need black-box prediction models learned from data, but also need explanation models that can reveal the physical insights underlying the data and model in human understandable ways, so that human beings can keep up with the pace of AI's discoveries.

Our experiments involve two types of models. One is a black-box model implemented as neural network (NN) which is learned from observational data. The black-box model is responsible for making accurate predictions such as predicting the position of Mars at certain time, as well as data augmentation to turn limited observational data into large scale data for science discovery.
The black-box model would have already been very helpful to human-beings, for example, it may help to develop calendars and to guide agricultural production by making accurate predictions of the future, but its black-box nature makes it difficult for humans to understand the underlying physical mechanism of such predictions. As a result, we involve the second type of model for explanation, which is implemented based on symbolic regression that transforms the black-box model into a symbolic function to express the interpretable physical rules. The symbolic regression process also discovers meaningful physical variables to inspire insightful understandings of the underlying physical mechanism behind the data.


For the implementation details, we use three layers of multi-layer perceptron (MLP) as the NN model with the hidden size as 100. 
As a black-box, we will not change the internal structure of NN throughout the experiments, i.e., we will not purposely design unique NN structures to fit different data, instead, we always use the same and simple three-layer MLP as the black-box and we only designate the input and output data for the black-box to learn different prediction models.
We use TuringBot \cite{2020turingBot} for model explanation based on symbolic regression, 
which is a widely used symbolic regression algorithm based on simulated annealing and performed well on a variety of physics-inspired learning problems \cite{ashok2020logic}.


\section{Rediscover Kepler's Laws based on Explainable AI}
\label{sec:rediscover_kepler_1}

Let us first review the process of Kepler's discovery of his first law. In Kepler's time, there were three models of planetary motion: the Ptolemaic, Copernican and Tychonic systems. In his book Astronomia Nova \cite{Astronomia}, Kepler mentioned that these three systems all had high prediction accuracy in the near term, but diverged and failed to fit historical and future observations in the long term. The first step of his research was to check the accuracy of the observation data. If a theory is based on inaccurate observations, then the theory could be misleading. Therefore, Kepler went through the calculation with at least seventy rounds of verification, at a very great loss of time \cite{koestler2017sleepwalkers}. Nowadays, data collection and inspection are still important and necessary but are relatively mature, and most part of them can be done automatically with minimal manual intervention. 

After multiple rounds of recalculation, Kepler chose to believe the observation data from Tycho. However, he was not satisfied with the measurement error of the existing planetary motion models \cite{caspar2012kepler}, which led him to propose a new hypothesis that the orbit of a planet is an ellipse with the Sun at one of the two foci, which is known as Kepler's first law of planetary motion, and then he used the observation data to verify his hypothesis. This process practiced the traditional Hypothetico-Deductive paradigm of science discovery, where a hypothesis is first manually proposed and then experiments are conducted to verify or falsify the hypothesis. In the following, we will show the hypothesis-free science discovery process based on (Explainable) AI which directly starts from data to rediscover the Kepler's laws.

\subsection{Black-box Model for Prediction and Data Augmentation}

\begin{figure*}[ht]
\begin{minipage}{0.5\textwidth}
  \centering
  \includegraphics[width=\textwidth]{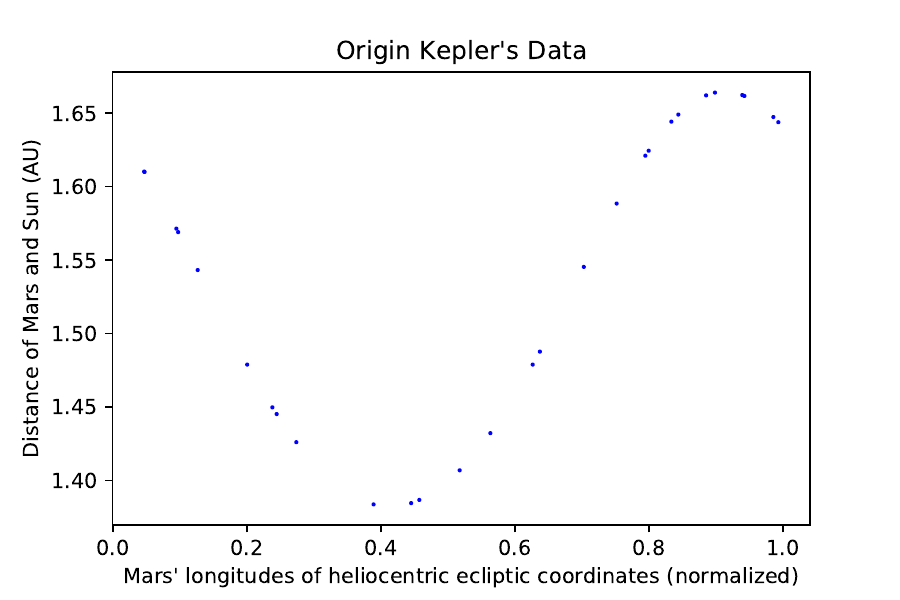}
  \vspace{-4ex}
  \caption{Data Visualization before Training}
  \label{fig:Kepler's original data}
\end{minipage}\hfill
\begin{minipage}{0.5\textwidth}
  \centering
  \includegraphics[width=\textwidth]{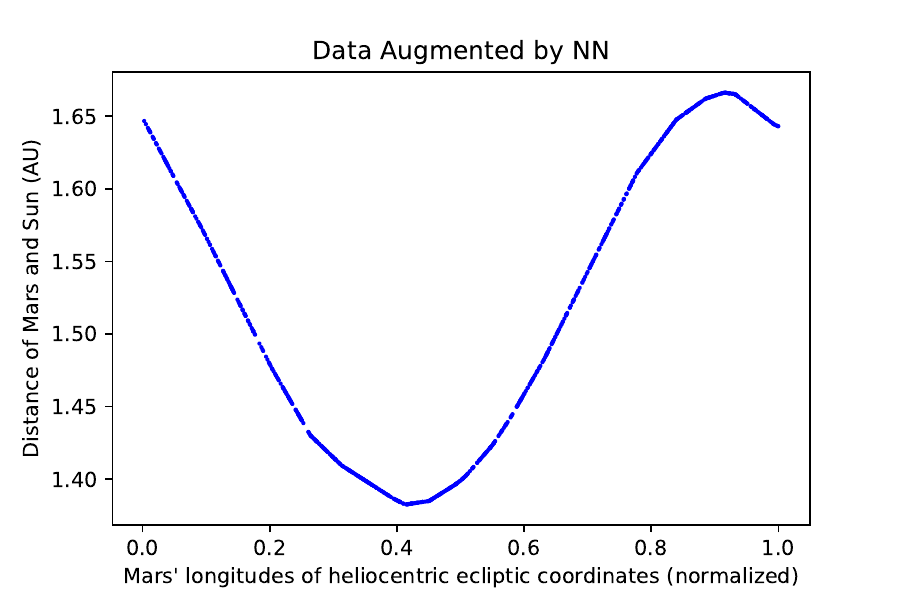}
  \vspace{-4ex}
  \caption{Data Visualization after Training}
  \label{fig:kepler's law data augmented by NN}
\end{minipage}
\end{figure*}

First, we use the neural network model as a black-box model for data fitting, prediction, and data augmentation. An advantage of deep neural network is its ability to smoothly fit the data so as to augment the small amount of observation data into large amount of data samples to facilitate AI-based science discovery.
We plot the observation data points in Table \ref{tab:position of Mars} as Figure \ref{fig:Kepler's original data}.
Since the amount of original observation data is small, we only use three samples for validation and use the remaining 25 samples for training. We set the number of training epochs as 200,000 for NN-based data fitting to learn a regression function $r=\text{NN}(\theta)$, where NN is the learned neural network function, $r$ is the Sun-Mars distance and $\theta$ is the Mars' angular position relative to the Sun (column 2 and column 3 of Table \ref{tab:position of Mars}).
The final mean square loss (MSE) of the NN on the training data and validation data is $4\times10^{-11}$ and $7\times10^{-8}$, respectively, which means the neural network function is able to provide quite accurate predictions. For data augmentation, we uniformly sample 1,000 random numbers between 0 and 1 as input to the NN, which is in the same input number range of the NN model. We then use the sampled inputs and the corresponding outputs of the NN model to generate augmented data samples and to approximate the function. The augmented data samples are shown as the Figure \ref{fig:kepler's law data augmented by NN}.

\subsection{White-box Model for Explanation}

Our next step is to interpret the neural network function $r=\text{NN}(\theta)$ into a human understandable symbolic function based on symbolic regression so as to gain physical insights from the black-box model.
From Figure \ref{fig:kepler's law data augmented by NN}, we can see that the NN model has a good ability of smooth function approximation 
and we can see the periodicity of data from the figure. However, even though our 3-layer MLP model is very simple from the AI perspective, it is still a very complex non-linear nested matrix multiplication formula, which is difficult to understand its physical insights.
This is why we use symbolic regression as Explainable AI to transform the black-box model into simple and intuitive physical rules. In particular, we hope to turn the neural network function $r=\text{NN}(\theta)$ into an explicit symbolic function $r=f(\theta)$.
We use the cosine function ($\cos$) due to periodicity alongside with three other basic operations for symbolic regression: addition (+), multiplication ($\cdot$) and division ($/$) (subtraction can be expressed by adding a negative sign). 
The symbolic regression results are shown in Table \ref{tab:sym result first law}.



\begin{figure*}[ht]
  \centering
  \includegraphics[width=0.5\textwidth]{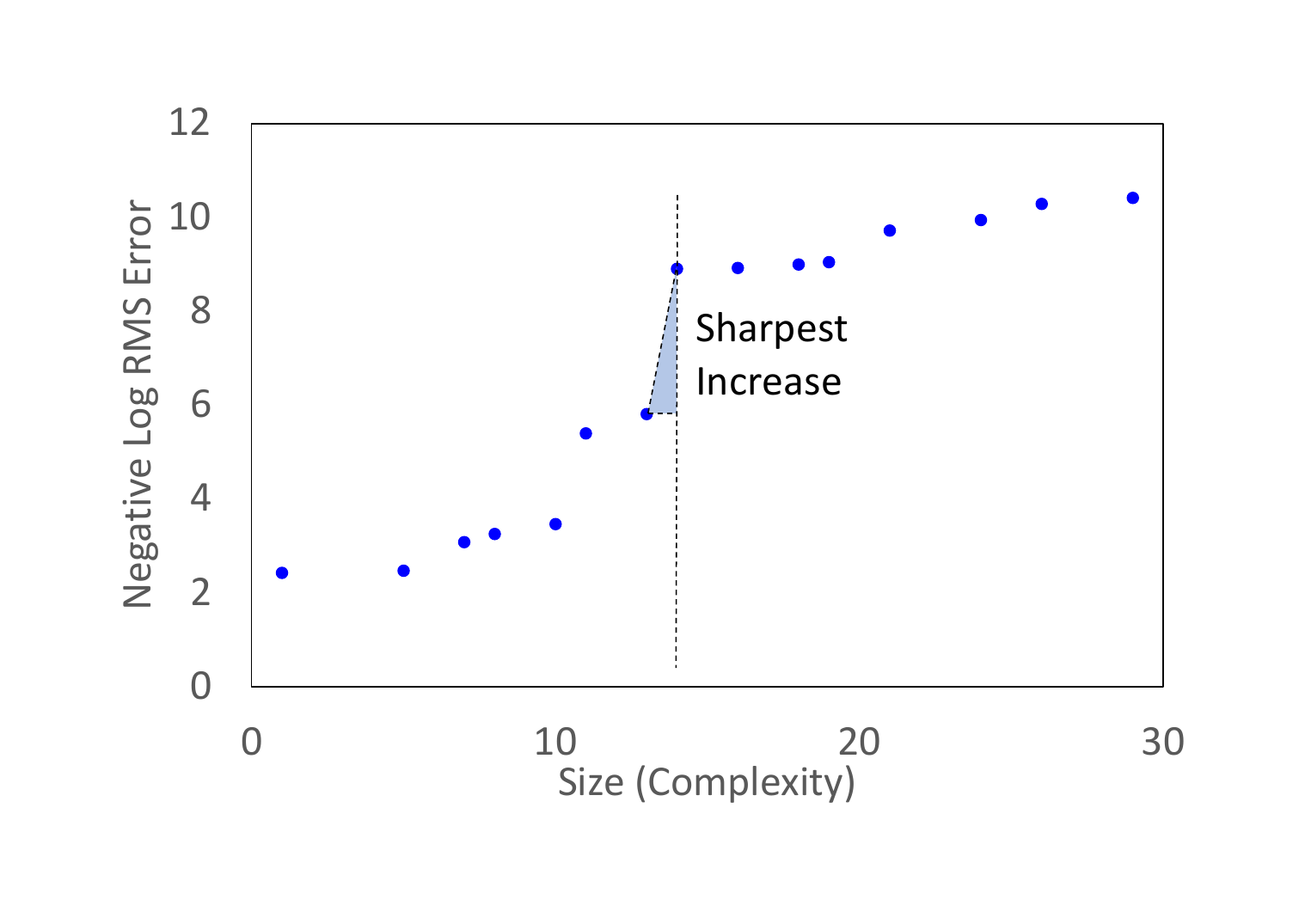}
  \vspace{-6ex}
  \caption{Size and Negative log Error}
  \label{fig:size and error result first law}
\end{figure*}

In Table \ref{tab:sym result first law}, the error means the root mean square error (RMSE) between the output of the black-box model and the output of the white-box model, i.e., $\text{RMSE}=\sqrt{\sum_\theta(\text{NN}(\theta)-f(\theta))^2}$. The size stands for the complexity of the generated function, which is calculated by adding up the size of each base function used by the generated function, and the size of each base function is shown in Table \ref{tab:size calculation}. The symbolic regression process selects the simple and effective functions, i.e., if a simpler (smaller size) function is more accurate (smaller error) than a complex (larger size) function, then the complex function will be eliminated from the results.
We plot the relation between the function size and the negative log error in Figure \ref{fig:size and error result first law}, which shows that
the size 14 candidate function has the sharpest increase in accuracy (in terms of negative log error) while maintaining a smaller size, which indicates that this function has the best chance to achieve a good balance between accuracy and complexity to reveal the physical rule behind the data \cite{udrescu2020ai}. We write down and simplify this function as Eq.\eqref{Eq:Kepler's first law}:
\begin{equation}
\begin{split}
    r = f(\theta)  &= \frac{1.51977}{1.00625 + 0.0932972\cdot \cos(\theta + 0.544536)} \\
    &= \frac{1.51033}{1 + 0.0927177\cdot \cos(\theta + 0.544536)}
\end{split}
    \label{Eq:Kepler's first law}
\end{equation}

If we have basic knowledge of elliptic equations, we know Eq.\eqref{Eq:Kepler's first law} implies a standard elliptical orbit, which can be represented as the following function in polar coordinates:
\begin{equation}
    r = f(\theta) = \frac{l}{1 + \varepsilon \cdot \cos(\theta)}
    \label{Eq:Standard oval equation}
\end{equation}
where $r$ stands for the distance between the Sun and Mars, and $\theta$ is Mars' longitude in heliocentric ecliptic coordinate. This clearly shows that the orbit of Mars is an ellipse with Sun at the focus, leading to Kepler's first law. We will further interpret the meaning of the numbers in Eq.\eqref{Eq:Kepler's first law} in the following subsection.

\subsection{Physical Interpretation of the Results}

Besides the elliptical orbit, we can obtain more insightful information from Eq.\eqref{Eq:Kepler's first law}. In Kepler's book Astronomia Nova \cite{Astronomia} (Chapter 41, Page 321), he calculated the eccentricity of Mars as $0.09264$ based on complex and meticulously designed geometric calculations. By comparing the two equations, Eq.\eqref{Eq:Kepler's first law} and Eq.\eqref{Eq:Standard oval equation}, we can directly learn that the Mars eccentricity $\varepsilon = 0.0927177$, which tends to be consistent with Kepler's result with relative error less than $0.1\%$. If we compare our Explainable AI-based result with the Mars eccentricity from modern science observations (as shown in Table \ref{tab:orbital characteristics of Mars}), we can see the relative error is about $0.7\%$, which is larger than that comparing with Kepler's result, most likely due to the observational errors in Tycho and Kepler's data that was collected 400 years ago, but the relative error is still small and the result is reasonable because we used the same data as Kepler did, and thus no surprisingly our result would be closer to Kepler's.

Another difference between Eq.\eqref{Eq:Kepler's first law} and the standard oval equation is the declination in the cosine function. For standard oval equation, $r_{min} = f(\theta = 0)$, while in our equation, $r_{min} = f[\theta = -0.544536\text{ (about}-31.2^\circ\text{)}]$. This is consistent with and can be explained by the series of closest Mars Oppositions in history. Mars Oppositions are phenomena when Earth passes in between Sun and Mars. Table \ref{tab:Mars Oppositions} (from \cite{meeus2003mars}) shows all of the closest Mars Oppositions with distance between Mars and Earth less than $0.375$ AU from the 1500s (Kepler's time) to nowadays. We see that all of these closest Mars Oppositions happen around August, which is about one month before the fall equinox (around September 23)---the time when Sun reaches the celestial longitude of $180^\circ$, i.e., $\theta_{Earth} = 0$. Since the orbit of Earth is very close to a circle (scientists at Kepler's time knew this according to their observations \cite{Astronomia}[p.271-272]), therefore, the distance between Mars and Earth on Mars Oppositions is mainly decided by the position of Mars, and Mars is most likely at perihelion around August according to the historical observations of closest Mars Oppositions (e.g., Table \ref{tab:Mars Oppositions}). This is consistent with the results shown by Explainable-AI model, since Eq.\eqref{Eq:Kepler's first law} also shows August as the Mars perihelion when $\theta_{Earth} = \theta_{Mars} = -0.544536 \approx -31.2^\circ$, which is about $\frac{31.2}{360}\times365\approx32$ days ahead of the fall equinox, i.e., in August. In the following, we will also show how Kepler's other laws can be discovered in the process of pursuing for Newton's laws based on Explainable AI.


\section{Rediscover Newton's Laws based on Explainable AI}
\label{sec:rediscover_newton}

We have shown how Kepler's first law and certain attributes of Mars can be extracted by Explainable AI from data. But one may not be satisfied with this, because one may naturally want to know why Mars orbits in oval and what ``power'' drives this elliptical orbit. In history, with limited information and tools, Kepler thought that the variable distance between Sun and Mars was due to the magnetic attraction and repulsion of Mars by Sun \cite{wilson1968kepler}, which was inspired from the proposal that Earth is a magnet by English physician and physicist William Gilbert in his groundbreaking book \textit{De Magnete} published in early 1600s  \cite{gilbert1958magnete}. Though we now know that this explanation is not the true reason, we cannot help imaging whether Explainable AI can help to answer this question based on ancient data that Kepler used.

\subsection{Black-box Model for Time-Sensitive Prediction and Data Augmentation}

\begin{figure*}[ht]
\begin{minipage}{0.5\textwidth}
  \centering
  \includegraphics[width=\textwidth]{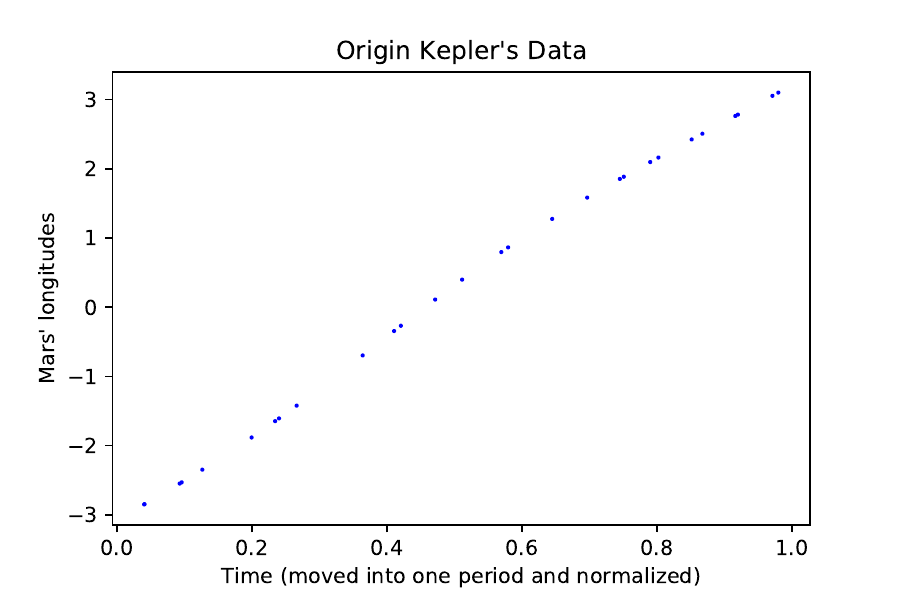}
  \vspace{-4ex}
  \caption{Data Visualization before Training}
  \label{fig:Kepler's original time position data}
\end{minipage}\hfill
\begin{minipage}{0.5\textwidth}
  \centering
  \includegraphics[width=\textwidth]{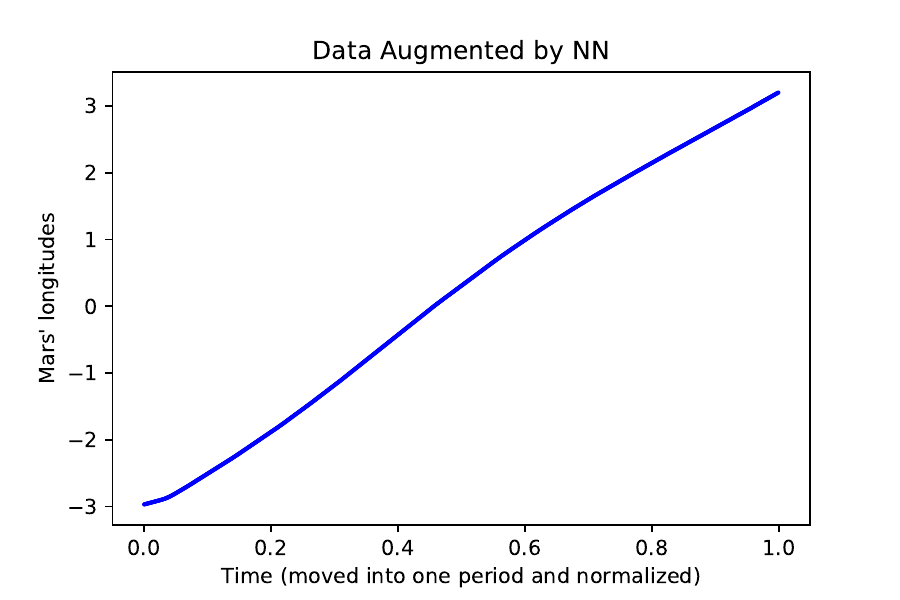}
  \vspace{-4ex}
  \caption{Data Visualization after Training}
  \label{fig:time position data augmented by NN}
\end{minipage}
\vspace{-3ex}
\end{figure*}

Eq.\eqref{Eq:Kepler's first law} describes the position of Mars relative to Sun as $r=f(\theta)$. An intuitive and interesting idea is to construct the relationship between the Mars position and time, since we have not used the time information in Table \ref{tab:position of Mars}. More specifically, we naturally hope to have the $\theta$-as-$t$ relationship $\theta=g(t)$, so that combined with the $r=f(\theta)$ function, we will be able to predict the Mars position for any given time in the future, which was an important problem for astronomy and calendar development at Kepler and Newton's time, and making accurate future predictions is also important to verify if a theory is correct. Kepler and his contemporary scientists knew that the orbital period of Mars is about 687 days, and the time span of the data in Table \ref{tab:position of Mars} is much longer than that, so we shift all data points into one orbital period and normalize the time to the range of [0,1] for 
better visualization. We show the normalized time $t$ and Mars' longitude $\theta$ in Figure \ref{fig:Kepler's original time position data}. We can see that the $\theta$-$t$ relationship is close to linear but with certain non-linearity, which implies small changes of Mars' speed when orbiting the Sun.

Similar to previous experiments, we first feed the data to a neural network model for black-box prediction and data augmentation. We use a simple three-layer multi-layer perceptron (MLP) network to train the predictor $\theta=\text{NN}(t)$, where the input is the normalized time $t$ and the output is Mars' longitude $\theta$ in radian. After $200,000$ epochs of training, the mean square loss (MSE) on training and validation data is $7\times10^{-8}$ and $1.5\times10^{-5}$, respectively. After training, we uniformly sample $2T$ points between 0 and 1 as input for data augmentation, 
where $T = 687$ is the orbital period of Mars, and we plot the augmented data points based on the trained neural predictor $\theta=\text{NN}(t)$ in Figure \ref{fig:time position data augmented by NN}.

Actually, the above simple experiment which adopts machine learning to learn a black-box neural predictor $\theta=\text{NN}(t)$ implies a significant role of machine learning (especially deep learning based on neural networks) in science discovery. Nowadays, based on advanced mathematical tools and deeper understandings of planetary motion, we are able to know that the relationship between $t$ and $\theta$ can be expressed as the following Eq.\eqref{Eq:function of t and theta} \cite{curtis2013orbital}, 
\begin{equation}
    \frac{2\pi}{T}t = 2\tan^{-1}\bigg(\sqrt{\frac{1-\epsilon}{1+\epsilon}}\tan\Big(\frac{\theta}{2}\Big)\bigg)-\frac{\epsilon\sqrt{1-\epsilon^2}\sin(\theta)}{1+\epsilon\cos(\theta)}
    \label{Eq:function of t and theta}
\end{equation}
where $T$ and $\epsilon$ are constant parameters of Mars. This means that we can express $t$ as a function of $\theta$, i.e., $t = h(\theta)$, however, we can hardly find a function to express $\theta$ as $t$, i.e., $\theta = g(t)$, since Eq.\eqref{Eq:function of t and theta} is a transcendental equation. As a result, suppose we did not know Eq.\eqref{Eq:function of t and theta}, then we possibly will spend efforts trying to find the $\theta$-$t$ relationship, however, any attempt to find a $\theta$-$t$ function $\theta=g(t)$ would fail no matter based on manual efforts or automatic tools such as symbolic regression, which incurs a waste of time. Nevertheless, sometimes we do need a $\theta$-as-$t$ function because we may want to analyze some important features of Mars motion such as the angular velocity and acceleration. Deep learning and neural network models provide a solution to this problem, because according to the universal approximation theorem \cite{cybenko1989approximation,hornik1991approximation,csaji2001approximation}, neural networks---when the structure and weights are properly designed and learned---are able to approximate a wide scope of functions based on training data. As a result, even though the functional form of $\theta = g(t)$ is difficult (if at all possible) to find, we can still learn a fairly good $\theta$-$t$ function as a neural network $\theta=\text{NN}(t)$, and because NN is differentiable, we can conduct mathematical analysis for the $\theta$-$t$ relationship, such as angular velocity $\omega=\frac{\text{dNN}(t)}{\text{d}t}$ and angular acceleration $a=\frac{\text{d}^2\text{NN}(t)}{\text{d}t^2}$. This will enable us to discover the possible physical relationship between many physical variables that are otherwise difficult to calculate. We will further illustrate on this point in the following subsections.


\subsection{White-box Model for Explanation}




At Kepler's time, calculus has not been invented yet, but scientists have already established the concept of velocity. As a result, we should not use differentials to derive the angular velocity, but we can calculate the angular velocity $\omega$ by calculating the change of angle $\theta$ within a certain time interval based on the black-box neural network prediction model $\theta=\text{NN}(t)$ that we have learned in the above subsection. 
To show the generalization ability of the neural network model, we randomly sample the same number (28) of data points from $\theta=\text{NN}(t)$ by using different values between 0.1 and 0.9 as the time in one orbital period of Mars, denoted as $t_i~(i=1,2,\cdots,28)$. For each $t_i$, we use the neural model to obtain the angle of Mars $\theta_i=\text{NN}(t_i)$. Then, we set the time interval $\delta_t = \frac{1}{32}$ day (about 45 minutes), and for each time $t_i$, we calculate the corresponding angular velocity $\omega_i = \frac{\text{NN}(t_i + \delta_t) - \text{NN}(t_i - \delta_t)}{2\delta_t}$. 
Besides, based on Eq.\eqref{Eq:Kepler's first law}, we can also obtain the corresponding Sun-Mars distance at time $t_i$ which is $r_i = f(\theta_i) =  f(\text{NN}(t_i))$.

Now, with the augmented data points $(t_i, \theta_i, r_i, \omega_i)$, we would like to find the interpretable symbolic physical relationship among some or all of these physical variables. Though nowadays even a high school student knows their relationship, we should assume that we have no knowledge about the underlying physical rule when using Explainable AI such as symbolic regression to find their relationship, because to demonstrate the role and ability of Explainable AI in this task, we should avoid from introducing prior knowledge into the discovery process. As a result, we augment as many variables as we can imagine from the existing variables and then totally rely on symbolic regression to find the potential relationship underlying some or all of the variables, including both existing variables and augmented variables.

\begin{figure*}[ht]
  \centering
  \includegraphics[width=0.6\textwidth]{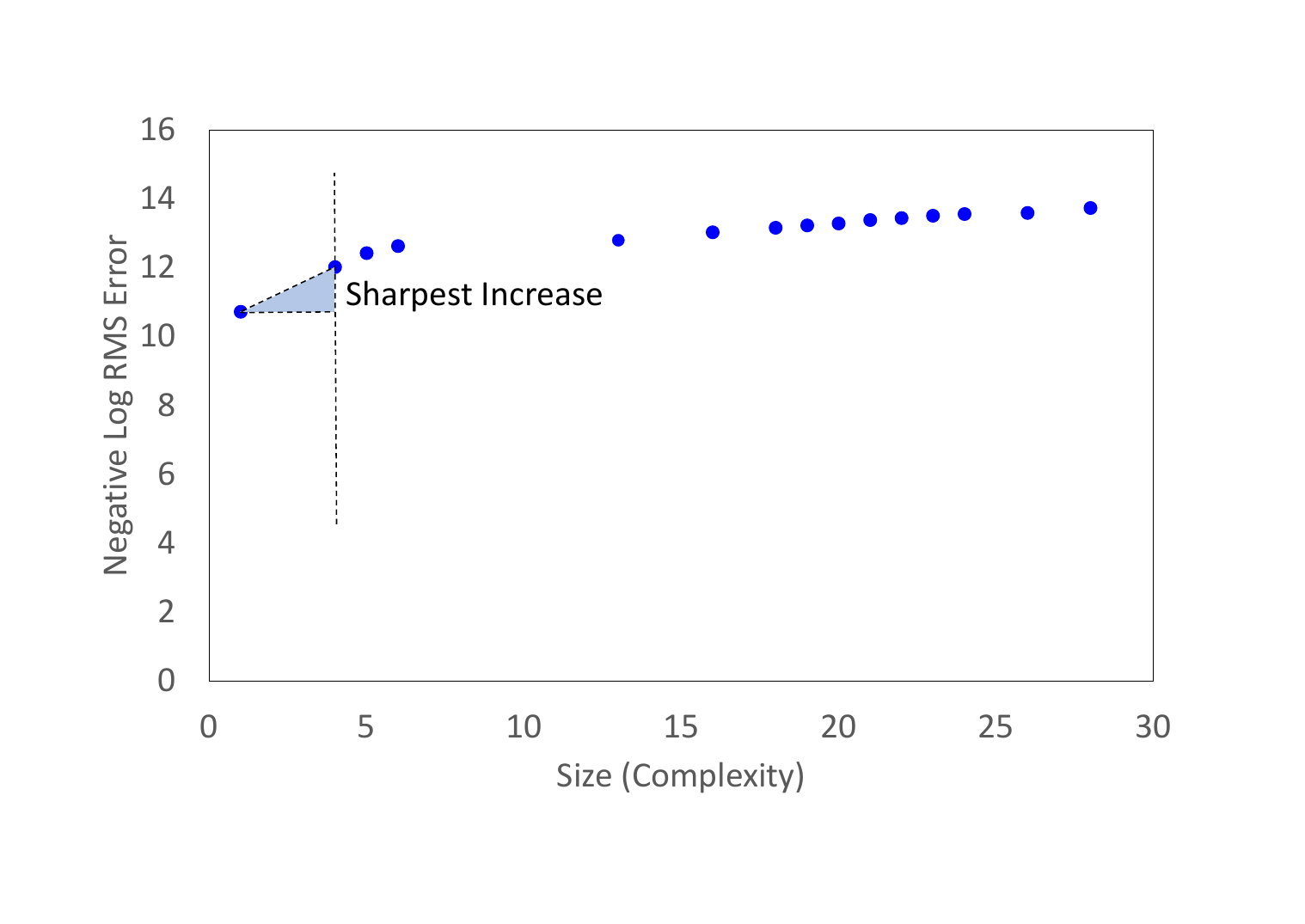}
  \vspace{-8ex}
  \caption{Size and Negative log Error for $r$ and $\omega$}
  \label{fig:size and error result Newton law}
  \vspace{-3ex}
\end{figure*}

As a demonstration to this process, we try to find the relationship between $r$ and $\omega$ (we can apply the same process on other pairs of variables if we want), and to reduce the complexity of the power operation, we create some augmented variables including $r_2=r^2, r_3=r^3$ and $\omega_2=\omega^2, \omega_3=\omega^3$. Finally, we use $r_1=r, r_2=r^2, r_3=r^3$ as the input variables to symbolic regress $\omega_1=\omega$, $\omega_2=\omega^2$ and $\omega_3=\omega^3$, respectively, and then, we repeat the process the other way around, i.e., use $\omega_1=\omega, \omega_2=\omega^2, \omega_3=\omega^3$ as the input variables to symbolic regress $r_1=r$, $r_2=r^2$, and $r_3=r^3$, respectively.
We still use four basic operations cosine ($\cos$), addition (+), multiplication ($\cdot$) and division ($/$) for symbolic regression, same as previous experiments. In this process, the symbolic regression results when finding function for $\omega_2 = F(r_1, r_2, r_3)$ are shown as Table \ref{tab:sym result Newton law}, and the relationship between function size and regression error (in terms of negative log RMSE) is in Figure \ref{fig:size and error result Newton law}.


From Figure \ref{fig:size and error result Newton law} we can see that the function with size 4 has the sharpest increase in accuracy, which indicates that this function has the best chance to achieve a good balance between accuracy and complexity to reveal the physical rule behind data \cite{udrescu2020ai}. We write down the function as follows (recall that $\omega_2=\omega^2, r_3=r^3$):
\begin{equation}
\label{eq:rw}
    \omega^2 = \frac{0.000298491}{r^3},~~~\text{or}~~~r^3\omega^2 = c = 0.000298491  AU^3day^{-2}
\end{equation}
Based on modern scientific knowledge, we know that $r^3\omega^2 = GM$, where $G=6.674\times10^{-11} m^3kg^{-1}s^{-2}$ is the gravitational constant and $M = 1.989 \times 10^{30} kg$ is the mass of Sun. Besides, our unit for distance is $1 AU = 1.496 \times 10^{11}m$, so we have $r^3\omega^2 = GM = 2.96 \times 10^{-4} AU^3day^{-2}$, and this number is very close to the constant in Eq.\eqref{eq:rw}, with relative error of about $0.8\%$, which is small and reasonable considering that we only used the ancient data that was collected 400 years ago. This shows the nice ability of Explainable AI such as symbolic regression in discovering the physical rules underlying data.

\subsection{Physical Interpretation of the Results}


Though we have shown that (Explainable) AI is able to make accurate predictions and generate explainable equations, we still want to emphasize that AI algorithms and machines do not ``understand'' or produce ``meaning'' for the physical variables or rules that emerge in the science discovery process.\footnote{We acknowledge that there exist debates over whether machines have their own internal ``meanings'' that are unknown or not understandable to humans, however, this is not a focus of this paper since we aim at science discovery for human beings.}
From the AI and machines' perspective, the variables and rules are just symbols and equations, and AI algorithms are only responsible for data analyses so as to extract possibly inspiring variables and rules, while it is human being's role to understand them and give meanings to the extracted variables and rules. For example, AI algorithms may discover a new combination of variables that as a whole is particularly useful for predicting and explaining the data (we will discuss with more details in the following). In this case, human experts may try to interpret the physical meaning of this combination of variables and leverage it to gain better understandings of the problem.
Sometimes, human experts may also need to innovatively create new physical concepts up front or with the assistance of AI during the discovery process so as to better interpret the results, enabling a human-AI collaborative science discovery process.

In the above experiment, once Explainable AI has generated the equation $r^3\omega^2 = c$ ($c$ is a constant), 
we first need to realize that $a = r\omega^2$ means the centripetal acceleration in circular motion. The deduction of centripetal acceleration $a$ does not need calculus, but acceleration was still a new concept at Kepler's time that needs to be created. With this new concept, $r^3\omega^2 = c$ can be reorganized into $ar^2 = c$, i.e., $a\propto\frac{1}{r^2}$, which is the centripetal acceleration equation. The centripetal acceleration equation $a\propto\frac{1}{r^2}$ can explain why the orbit of Mars is elliptical, but to really know what causes the elliptical orbit, we still need to know what causes the centripetal acceleration $a$, i.e., what is the underlying ``force'' that drives the Mars orbiting in such a way. 

At Kepler's time, scientists already had the concept of force, but they did not know the correct function of force, since they thought force was proportional to distance and thus speed. It was Newton's greatness to realize that force is not the reason of speed but actually the reason of acceleration, and he innovatively connected force with acceleration by $F = ma$. Based on this, suppose $M$ is the mass of Sun, $ar^2 = c$ can be reorganized as $Mar^2 = Mc$, thus $Fr^2 = Mc$, and $F \propto \frac{1}{r^2}$, leading to Newton's inverse-square law of universal gravitation. One can see that in this interpretation process, humans still need to play an important role in understanding or creating physical concepts. This is partly because at Kepler and Newton's time, measuring the force is almost impossible, and thus the observational data does not include force $F$ as one of the variables. If force $F$ were one of the variables with observed values, just like time $t$ and angle $\theta$, then (Explainable) AI methods might be able to extract the symbolic equation for $F$ from data, just as we did in the above for other variables, which can save efforts for human exploration, intervention, interpretation and creation in the discovery process. However, even if we have observed $F$ values, like in many modern scientific datasets, we can never fully exclude the possibility that human experts need to innovatively create other new concepts that do not yet exist in the observational data.


\subsection{Relation with Kepler's Third Law}


The discovered rule $r^3\omega^2 =c$ in Eq.\eqref{eq:rw} could be mis-interpreted as the Kepler's Third Law $\frac{\bar{r}^3}{T^2} =c^\prime$ if one 
applies $\bar{\omega}=\frac{2\pi}{T}$ for circular motion. Actually, Eq.\eqref{eq:rw} should not be interpreted as the Kepler's Third Law, because Kepler's Third Law talks about a universal rule for all planets circulating Sun, while Eq.\eqref{eq:rw} only talks about a rule for Mars since this rule is solely derived from Mars data. More specifically, Eq.\eqref{eq:rw} is saying that at any time $t$, the Mars' angular velocity $\omega$ and its corresponding distance to Sun $r$ satisfy a constant rule, while Kepler's Third Law is saying that for all planets circulating Sun, their mean distance to Sun and circulating period satisfy a constant rule. To really discovery and justify Kepler's Third Law, we need to include the data of more planets. Actually, Kepler himself studied six planets: Mercury, Venus, Earth, Mars, Jupiter and Saturn. This example shows that we need to be extremely careful when trying to interpret the Explainable AI discovered rules and
avoid from over-generalizing the results.

However, Eq.\eqref{eq:rw} is still useful and may inspire us towards Kepler's Third Law. If we look at this equation from a macro perspective, the period $T$ can be considered as an integrated effect of the angular velocity $\omega$, which may lead us to consider whether $T$ and $r$ also exhibit similar rules. Even though we know that the Kepler's Third Law $\frac{\bar{r}^3}{T^2} = c^\prime$ needs information of other planets to justify, Eq.\eqref{eq:rw} which is solely derived from the Mars data may point to the right direction and speed up the discovery process.
Actually, if we take $\bar{\omega}=\frac{2\pi}{T}$ into Eq.\eqref{eq:rw}, we have $\frac{\bar{r}^3}{T^2} = \frac{c}{4\pi^2} = 7.56086\times 10^{-6} AU^3day^{-2} \doteq c^\prime$, which is close Kepler's result ($7.5\times10^{-6} AU^3day^{-2}$, error within 0.82\%) and modern science result ($7.495\times10^{-6} AU^3day^{-2}$, error within 0.88\%). 
The ability of associating macro and micro perspectives by intuition is uniquely owned by humans instead of current AI, and this is one of the reasons why humans still play an indispensable role in modern science discovery process.




\section{Conclusions and Future Work}
\label{sec:conclusions}

In this paper, we highlight the role of Explainable AI in science discovery by demonstrating an Explainable AI-based paradigm for science discovery. To demonstrate the idea, we show how Kepler's laws of planetary motion and Newton's law of universal gravitation can be rediscovered with the assistance of Explainable AI based on a small amount of Tycho Brahe's astronomical observation data, whose works were leading the scientific revolution in the 16-17th century. Technically, we use black-box models such as deep neural networks for prediction and data augmentation, and use white-box models such as symbolic regression for model explanation. Insightful discoveries and conclusions can be derived by interpreting the results under the assistance of Explainable AI. We also demonstrate the indispensable role of human beings in the science discovery process on creating new concepts, assigning meanings to the discovered variables and rules, and providing insightful intuitions to supervise the AI-based discovery process.

In the future, we will further refine the Explainable AI-based paradigm for science discovery by considering a wider scope of black-box and white-box models as well as applying them to various different scientific fields. 
We will also use the framework for more state-of-the-art scientific problems such as dark matters based on modern astronomical observation data and particle physics based on the data collected from Large Hadron Colliders for discovering new knowledge that is unknown to human.
And probably one day, we can even improve the performance of Explainable AI through the knowledge discovered by AI itself while maintaining complete control of the process by demanding explanations from AI, since we always stand on the shoulder of giants.

\bibliographystyle{unsrt}
\bibliography{references}






\newpage

\onecolumn

\section{Appendix}

In this section, we present all of the tables and figures referenced in the paper.


\begin{table}[ht]
    \centering
    \begin{tabular}{c|c|c|c}
        \toprule
        Time (YYYY/MM/DD) & Mars' Position in Ecliptic & Sun-Mars Distance & Difference \\
        \midrule
        1582/11/23 16:00 &  90.70306$^\circ$ & 1.58852 & $+1^\prime30^{\prime \prime}$ \\
        1582/12/26 08:30 & 106.12167$^\circ$ & 1.62104 & $+3^\prime49^{\prime \prime}$ \\
        1582/12/30 08:10 & 107.94222$^\circ$ & 1.62443 & $+5^\prime50^{\prime \prime}$ \\
        1583/01/26 06:15 & 120.10667$^\circ$ & 1.64421 & $-2^\prime33^{\prime \prime}$ \\
        \midrule
        1584/12/21 14:00 & 123.86250$^\circ$ & 1.64907 & $+1^\prime04^{\prime \prime}$ \\
        1585/01/24 09:00 & 138.78556$^\circ$ & 1.66210 & $-3^\prime32^{\prime \prime}$ \\
        1585/02/04 06:40 & 143.56139$^\circ$ & 1.66400 & $-3^\prime08^{\prime \prime}$ \\
        1585/03/12 10:30 & 159.38722$^\circ$ & 1.66170 & $-2^\prime29^{\prime \prime}$ \\
        \midrule
        1587/01/25 17:00 & 158.22778$^\circ$ & 1.66232 & $-0^\prime10^{\prime \prime}$ \\
        1587/03/04 13:24 & 174.94722$^\circ$ & 1.64737 & $-0^\prime59^{\prime \prime}$ \\
        1587/03/10 11:30 & 177.59833$^\circ$ & 1.64382 & $0^\prime0^{\prime \prime}$ \\
        1587/04/21 09:30 & 196.74750$^\circ$ & 1.61027 & $+1^\prime30^{\prime \prime}$ \\
        \midrule
        1589/05/08 16:24 & 196.92056$^\circ$ & 1.61000 & $-2^\prime43^{\prime \prime}$ \\
        1589/04/13 11:15 & 214.03056$^\circ$ & 1.57141 & $+1^\prime40^{\prime \prime}$ \\
        1589/04/15 12:05 & 215.02806$^\circ$ & 1.56900 & $+0^\prime37^{\prime \prime}$ \\
        1589/05/06 11:20 & 225.51000$^\circ$ & 1.54326 & $+0^\prime57^{\prime \prime}$ \\
        \midrule
        1591/05/13 14:00 & 252.12722$^\circ$ & 1.47891 & $-4^\prime24^{\prime \prime}$ \\
        1591/06/06 12:20 & 265.64667$^\circ$ & 1.44981 & $-3^\prime15^{\prime \prime}$ \\
        1591/06/10 11:50 & 267.94694$^\circ$ & 1.44526 & $-4^\prime39^{\prime \prime}$ \\
        1591/06/28 10:24 & 278.49222$^\circ$ & 1.42608 & $-5^\prime39^{\prime \prime}$ \\
        \midrule
        1593/07/21 14:00 & 320.02722$^\circ$ & 1.38376 & $-2^\prime31^{\prime \prime}$ \\
        1593/08/22 12:20 & 340.25694$^\circ$ & 1.38463 & $-0^\prime36^{\prime \prime}$ \\
        1593/08/29 10:20 & 344.62083$^\circ$ & 1.38682 & $-2^\prime19^{\prime \prime}$ \\
        1593/10/03 08:00 &   6.32750$^\circ$ & 1.40697 & $-0^\prime16^{\prime \prime}$ \\
        \midrule
        1595/09/17 16:45 &  22.82194$^\circ$ & 1.43222 & $-1^\prime27^{\prime \prime}$ \\
        1595/10/27 12:20 &  45.59389$^\circ$ & 1.47890 & $-0^\prime29^{\prime \prime}$ \\
        1595/11/03 12:00 &  49.44250$^\circ$ & 1.48773 & $+0^\prime03^{\prime \prime}$ \\
        1595/12/18 08:00 &  73.04139$^\circ$ & 1.54539 & $-0^\prime59^{\prime \prime}$ \\
        \bottomrule
    \end{tabular}
    \caption{Position of Mars when orbiting the Sun}
    \label{tab:position of Mars}
    \vspace{-3ex}
\end{table}

\begin{table*}[ht]
    \centering
    \begin{tabular}{l|c|l}
        \toprule
         Size & Error & Function \\
         \midrule
         1 & 0.088419 & $1.54806$ \\
         5 & 0.084370 & $1.54329+0.0130577\cdot\theta$ \\
         7 & 0.045791 & $1.45537+0.021878\cdot\theta\cdot\theta$ \\
         8 & 0.038594 & $1.53256-0.101048\cdot\cos{\theta}$ \\
         10 & 0.031201 & $1.65411-\frac{0.321963}{1.21921+\theta\cdot\theta}$ \\
         11 & 0.004519 & $1.51578-0.142019\cdot\cos{(\theta+0.542453)}$ \\
         13 & 0.003003 & $1.51836 - 0.141285\cdot\cos{(0.979081\cdot(-0.544189-\theta))}$ \\
         14 & 0.000136 & $\frac{1.51977}{1.00625+0.0932972\cdot\cos(\theta+0.544536)}$ \\
         16 & 0.000133 & $\frac{1.51975}{1.00625+0.0933058\cdot\cos(1.00017\cdot\theta+0.544619)}$ \\
         18 & 0.000124 & $\frac{1.5221}{1.0078+\sin(0.0935495\cdot\cos(\theta+0.544689))}$ \\
         19 & 0.000118 & $1.51016 - \frac{0.0794197}{0.0536393+\frac{0.567314}{\cos(1.00052\cdot(0.544488+\theta))}}$ \\
         21 & 0.000060 & $\frac{1.51978}{1.00625+0.0932649\cdot\cos(\theta + 0.544414 + \frac{0.000322752}{\theta-1.48167})}$ \\
         24 & 0.000048 & $1.51031-\frac{0.0793261}{0.052716+\frac{0.56737}{\cos(0.543701+\theta)}-\frac{0.000771507}{1.48757-\theta}}$ \\
         26 & 0.000034 & $1.51023 - \frac{0.0793521}{0.0531939+\frac{0.56753}{\cos(0.543898+1.00028\cdot\theta)}-\frac{0.00067777}{1.49235-\theta}}$ \\
         29 & 0.000030 & $1.51032 - \frac{\cos(-0.543588-\theta)}{7.14743+0.668919\cdot\cos(0.55992-\frac{0.00769976}{1.58368-\theta} + \theta)}$ \\
         \bottomrule
    \end{tabular}
    \caption{Symbolic Regression Results}
    \label{tab:sym result first law}
    \vspace{-3ex}
\end{table*}

\begin{table*}[ht]
    \centering
    \begin{tabular}{c|l}
        \toprule
         Size & Base functions \\
         \midrule
         Size 1 & an input variable, addition (+), subtraction (-), and multiplication ($\cdot$) \\
         Size 2 & division (/)\\
         Size 4 & other functions used in this paper \\
         \bottomrule
    \end{tabular}
    \caption{Based Functions for Symbolic Regression and their Size}
    \label{tab:size calculation}
\end{table*}

\begin{table*}[ht]
    \centering
    \begin{tabular}{c|l|l}
        \toprule
         Parameter & Description & Value \\
         \midrule
         $\varepsilon$ & Eccentricity & 0.093 412 33 \\
         $a$ & Semi-major axis & 1.523 662 31 AU \\
         $T$ & Orbital period & 686.971 days \\
         \bottomrule
    \end{tabular}
    \caption{Orbital Parameters of Mars based on Modern Science}
    \label{tab:orbital characteristics of Mars}
\end{table*}

\begin{table*}[ht]
    \centering
    \begin{tabular}{c|c|c}
        \toprule
         Year (AD) & Date & Earth-Mars Distance in AU \\
         \midrule
         1561 & Aug. 07 & 0.37325 \\
         1640 & Aug. 20 & 0.37347 \\
         1687 & Aug. 09 & 0.37434 \\
         1719 & Aug. 25 & 0.37401 \\
         1766 & Aug. 13 & 0.37326 \\
         1845 & Aug. 18 & 0.37302 \\
         1924 & Aug. 22 & 0.37285 \\
         2003 & Aug. 27 & 0.37272 \\
         2050 & Aug. 15 & 0.37405 \\
         \bottomrule
    \end{tabular}
    \caption{Closest Approaches of Mars Oppositions in History}
    \label{tab:Mars Oppositions}
\end{table*}

\begin{table*}[ht]
    \centering
    \begin{tabular}{l|c|l}
        \toprule
         Size & Error & Function \\
         \midrule
         1 & 0.000022 & $8.18954\times 10^{-5}$ \\
         4 & 0.000006 & $\frac{0.000298491}{r_3}$ \\
         5 & 0.000004 & $0.000218591\cdot(1.92033-r)$ \\
         6 & 0.000003 & $-2.65592\times10^{-5}+\frac{0.000390417}{r_3}$ \\
         13 & 0.000003 & $-8.50685\times10^{-5}+\frac{0.000395123}{r_2-\frac{0.000290053}{r-1.48997}}$ \\
         16 & 0.000002 & $\frac{0.000100316}{-1.08788-0.0590273\cdot\cos(-2147483648\cdot r_3)+r_2)}$ \\
         22 & 0.000001 & $\frac{0.000448514}{(\frac{0.0460772}{r_3-3.36628}\cdot\cos(\frac{r_3}{-3.79879\times10^{-5}})+r_3)\cdot r}$ \\
         \bottomrule
    \end{tabular}
    \caption{Symbolic Regression Result for $r$ and $\omega$}
    \label{tab:sym result Newton law}
\end{table*}


\end{document}